\documentclass[10pt,twocolumn,letterpaper]{article}

\usepackage{cvpr}              

\usepackage[dvipsnames]{xcolor}

\definecolor{cvprblue}{rgb}{0.21,0.49,0.74}
\usepackage[pagebackref,breaklinks,colorlinks,citecolor=cvprblue]{hyperref}

\def\paperID{*****} 
\def\confName{CVPR}
\def\confYear{2024}

\title{Enhancing Subsequent Video Retrieval via Vision-Language Models (VLMs)}

\author{Yicheng Duan, Xi Huang, Duo Chen\\
Computer and Data Sciences, School of Engineering\\Case Western Reserve University\\
Cleveland, Ohio, 44106, USA\\ 
{\tt\small yxd245@case.edu}
{\tt\small xxh580@case.edu}
{\tt\small dxc830@case.edu}
}

\begin{document}
\maketitle
\begin{abstract}
The rapid growth of video content demands efficient and precise retrieval systems. While vision-language models (VLMs) excel in representation learning, they often struggle with adaptive, time-sensitive video retrieval. This paper introduces a novel framework that combines vector similarity search with graph-based data structures. By leveraging VLM embeddings for initial retrieval and modeling contextual relationships among video segments, our approach enables adaptive query refinement and improves retrieval accuracy. Experiments demonstrate its precision, scalability, and robustness, offering an effective solution for interactive video retrieval in dynamic environments.
\end{abstract}    
\section{Introduction}
\label{sec:intro}
The exponential growth of multimedia content, especially TV shows and news videos, has created a pressing demand for efficient video retrieval systems. Traditional methods often fail to capture temporal dependencies and contextual relationships in long-duration videos, limiting their ability to retrieve meaningful subsequent video segments based on a given frame or query.

To address these challenges, we propose a framework that combines Vision-Language Models (VLMs) with graph-based structures for enhanced video retrieval. Using the Redhen TV Show dataset, which includes diverse video formats and metadata like subtitles and timestamps, our method achieves high-quality temporal alignment and retrieval accuracy.

Our contributions include: (1) developing a prompt engineering strategy to improve VLM embeddings; (2) integrating Pinecone for vector similarity search with Neo4j for managing graph-structured metadata; and (3) optimizing retrieval processes for multilingual and cross-domain scenarios. Experimental results demonstrate the system's ability to retrieve and summarize relevant video clips effectively, showcasing its robustness and scalability in diverse scenarios.

\section{Related work}
\label{sec:related_work}

The task of video retrieval has progressed significantly through three key stages: traditional methods, deep learning-based models, and the advent of large language models (LLMs) and vision-language models (VLMs). Each stage brought significant advancements, backed by foundational research.

Early video retrieval approaches relied on handcrafted features and classical algorithms, which, despite their simplicity, laid the groundwork for subsequent innovations. Collaborative filtering (CF) was one of the earliest and most successful methods in recommender systems. Sarwar et al. proposed item-based CF to improve scalability and efficiency in retrieval tasks\cite{10.1145/371920.372071}. While effective, these methods struggled to incorporate temporal information in video data.

The advent of deep learning enabled more advanced video retrieval by leveraging rich semantic representations and improved modeling of temporal relationships. Deep neural networks like the one proposed by Covington et al. in YouTube’s recommendation system used a two-tower architecture to separately learn embeddings for users and items. This approach proved highly scalable for large-scale video recommendation.\cite{10.1145/2959100.2959190} RNNs, including LSTMs and GRUs, were applied to model temporal dependencies in videos. Srivastava et al. demonstrated the use of sequence-to-sequence RNNs for video prediction tasks, laying the groundwork for their use in video retrieval.\cite{srivastava2016unsupervisedlearningvideorepresentations}

The most recent developments leverage multimodal pre-trained models to integrate visual and textual information, enabling more effective video retrieval. The paper "HourVideo: 1-Hour Video-Language Understanding" introduces a benchmark dataset for long-form video-language understanding, focusing on tasks such as summarization, perception, visual reasoning, and navigation across 500 curated egocentric videos. The study evaluates multimodal models like GPT-4 and LLaVA-NeXT, revealing a substantial performance gap between state-of-the-art models and human experts. For example, the Sequence Recall task evaluates a model's ability to memorize and retrieve specific event sequences from long videos. The task requires the model to understand dispersed temporal segments and correctly recall the sequence of relevant events based on given prompts. Experimental results reveal that current multimodal models significantly underperform compared to human experts, with models achieving an average accuracy of 25\%-35\%, while human performance exceeds 80\%, highlighting the challenges and limitations of existing models in long video understanding tasks.\cite{chandrasegaran2024hourvideo1hourvideolanguageunderstanding}
\section{Methods}
\label{sec:methods}

\subsection{Dataset}
The dataset we used is Redhen dataset (about 600k)\cite{redhen_dataset}. The Redhen Dataset is a comprehensive collection of television program recordings, organized by broadcast dates. It includes various data types, such as video files (MP4) with durations ranging from 10 minutes to an hour and subtitle files (TXT) with precise timestamps, making it a valuable resource for multimedia analysis and natural language processing research

\subsection{Vision-Language Model}
Our framework utilizes a Vision-Language Model (VLM) to generate embeddings for video frames, capitalizing on the VLM’s capabilities in instruction following and visual comprehension. For this task, we have selected the Qwen2-VL-2B-Instruct model as our primary VLM due to its robust performance in understanding and interpreting visual and textual information \cite{Qwen2VL}. 

\subsection{Video prepossessing Stage}
\subsubsection{Video segmentation and frame extraction}
We define a video segmentation interval \( V \), where every \( V \) seconds, a frame \( F_i \) is extracted from the video. Our transcription dataset \( C \) includes timestamps for each sentence. By aligning each frame \( F_i \) with its corresponding transcription \( C_i \) using these timestamps, we construct a paired dataset:
\[
\mathcal{D} = \left\{ (F_i, C_i) \right\}_{i=1}^{N}
\]
where \( N \) is the total number of extracted frames.

This dataset \( \mathcal{D} \) enables the association of visual content with textual descriptions
\subsubsection{Prompt Engineering Strategy}

To facilitate effective vector embedding via VLM using both images and transcribed text, we employed a prompt engineering strategy that leverages distinct system and user roles, supported by our Visual Language Model (VLM). The prompt structure is defined as follows:

\[
\text{Func\_prompt} =
\begin{cases}
\begin{aligned}
& \text{System:} \quad "\text{You are a retraval helper for  }\\ 
& \quad \text{generating better words(vectors)} \\
& \quad \text{for retriveing videos by given image} "\\
& \text{User:} \quad "\text{Image: }F_i ;\\
& \quad \text{Text: } C_i + C_{\text{BackgroundInfo}};"
\end{aligned}
\end{cases}
\]

Where:
\begin{itemize}
    \item \( F_i \) denotes the extracted video frame at the \( i \)-th segmentation interval \( V \).
    \item \( C_i \) represents the transcription corresponding to frame \( F_i \).
    \item \( C_{\text{BackgroundInfo}} \) includes additional contextual information relevant to the story.
\end{itemize}

We aggregate these prompt structures into a set \( \mathcal{D}_{\text{prompt}} \):

\[
\mathcal{D}_{\text{prompt}} = \left\{ \text{Func\_prompt}_i \right\}_{i=1}^{N}
\]

where \( N \) is the total number of prompt instances corresponding to the extracted frames.

This structured approach enables the VLM to seamlessly integrate visual and textual data, generating coherent and contextually rich story summaries. By aligning each frame \( F_i \) with its respective transcription \( C_i \) and additional background information \( C_{\text{BackgroundInfo}} \), the model can effectively interpret and summarize the narrative conveyed in the video.

\bigskip

\textbf{Example:} For a video segmented every \( V = 10 \) seconds, suppose we have:
\begin{itemize}
    \item Frame \( F_1 \) at 10 seconds.
    \item Transcription \( C_1 \) corresponding to \( F_1 \).
    \item Background information \( C_{\text{BackgroundInfo}} \) providing additional context.
\end{itemize}

The corresponding prompt \( \text{Func\_prompt}_1 \) would be:
\[
\text{Func\_prompt}_1 =
\begin{cases}
\begin{aligned}
& \text{System:} \quad "\text{You are a retraval helper for  }\\ 
& \quad \text{generating better words(vectors)} \\
& \quad \text{for retriveing videos by given image} "\\
& \text{User:} \quad "\text{Image: }F_1 ;\\
& \quad \text{Text: } C_1 + C_{\text{BackgroundInfo}};"
\end{aligned}
\end{cases}
\]

\subsubsection{VLM embedding}
Given \( \mathcal{D}_{\text{prompt}} \), each \( \text{Func\_prompt}_i \) is processed through the Visual Language Model (VLM) in a forward pass to generate embeddings for retrieval. We propose four methods:

\begin{enumerate}
    \item \textbf{Simple Mean Pooling:}
    Averages all token embeddings in the final hidden layer:
    \[
    \text{Embedding}_i = \frac{1}{T} \sum_{t=1}^{T} \text{HiddenState}_{t}
    \]

    \item \textbf{Max Pooling:}
    Selects the maximum value across token embeddings:
    \[
    \text{Embedding}_i = \max_{t=1}^{T} \text{HiddenState}_{t}
    \]

    \item \textbf{Mean Pooling with Attention Mask:}
    Averages embeddings while excluding padding tokens using an attention mask:
    \[
    \text{Embedding}_i = \frac{\sum_{t=1}^{T} \text{Mask}_t \cdot \text{HiddenState}_t}{\sum_{t=1}^{T} \text{Mask}_t}
    \]

    \item \textbf{Concatenation of Multiple Layers:}
    Combines multiple \(L\) hidden layers by averaging selected layers after mean pooling with an attention mask:
    \[
    \begin{aligned}
    &  \text{Embedding}_i = \\
    &   \text{Normalize}\left( \frac{1}{n} \sum_{l=L-n+1}^{L} \frac{\sum_{t=1}^{T} \text{Mask}_t \cdot \text{HiddenState}_{t, l}}{\sum_{t=1}^{T} \text{Mask}_t} \right)
    \end{aligned}
    \]
\end{enumerate}

Then, combining these embeddings (\( \text{vector}_i \)) with the earlier extracted video frame (\( F_i \)) and its corresponding timestamp (\( T_{v_i} \)), we construct a new dataset:
\[
\mathcal{D}_{\text{retrieval}} = \left\{ (F_i, T_{v_i}, \text{vector}_i) \right\}_{i=1}^{N}
\]
where \( N \) is the total number of frames. This dataset is foundational for subsequent retrieval tasks, enabling efficient and accurate video search by leveraging the combined power of visual, temporal, and semantic representations.

\subsubsection{Data Storage}

Efficient and scalable data storage is a critical component for supporting video frame retrieval in this project. To achieve this, we adopt a hybrid approach using Pinecone and Neo4j, each playing complementary roles in storing and retrieving video frame data.

\textbf{Pinecone}: A vector database designed for high-speed similarity searches and dense vector storage. It stores the feature vectors \( \text{vector}_i \) generated by the Vision-Language Model (VLM). These vectors enable efficient retrieval of the top-\( k \) similar items based on a query:
\[
\text{Pinecone\_result} = \{\text{vector}_{i_1}, \text{vector}_{i_2}, \dots, \text{vector}_{i_k}\}
\]

\textbf{Neo4j}: A graph database optimized for managing relationships and structured data. It is used to store and organize metadata \( \mathcal{M}_i \) associated with each frame \( F_i \), including:
\[
\mathcal{M}_i = \{ \text{FilePath}_i, \text{Subtitle}_i, T_{v_i} \}
\]
where \( T_{v_i} \) represents the timestamp of frame \( F_i \). In our design, frames are stored as nodes in a graph, and each node with an earlier timestamp \( T_{v_i} \) points to the node with a later timestamp \( T_{v_{i+1}} \), forming a directed linked list for each video:
\[
F_1 \xrightarrow{T_{v_1}} F_2 \xrightarrow{T_{v_2}} \cdots \xrightarrow{T_{v_N}}
\]

\textbf{Integration of Pinecone and Neo4j:} Pinecone vectors \( \text{vector}_i \) are ID-aligned with Neo4j nodes \( \text{Node}_i \). Retrieved top-\( k \) vector IDs query Neo4j for frame metadata \( \mathcal{M}_i \), ensuring seamless linkage between embeddings and metadata.

\textbf{Advantages of the Hybrid Design}: The combined use of Pinecone and Neo4j offers:
\begin{itemize}
    \item \textbf{Efficiency:} Pinecone supports fast and accurate vector similarity matching, even with large-scale datasets.
    \item \textbf{Flexibility:} Neo4j enables advanced navigation of frame sequences and supports complex contextual queries.
    \item \textbf{Scalability:} The hybrid design handles large volumes of video data while maintaining high retrieval performance.
\end{itemize}

By leveraging Pinecone for vector similarity searches and Neo4j for relational and temporal metadata storage, our system ensures efficient, flexible, and scalable video frame retrieval.

\subsection{Retrieval Stage}

\subsubsection{VLM Embedding for Input Image}

For input images, we use the Visual Language Model (VLM) to generate embeddings directly from the visual input \( F_\text{in} \). Unlike the approach in 3.2.2, this process does not incorporate any additional prompt text \( C_i + C_{\text{BackgroundInfo}} \). The output is a dense feature vector \( \text{V}_\text{in} \), representing the VLM's inherent semantic and visual characteristics of the input image. 

\subsubsection{Retrieval Method}

The retrieval process involves a seamless integration of Pinecone and Neo4j to handle vector matching and contextual metadata retrieval. The detailed steps are as follows:

\begin{enumerate}
    \item \textbf{Vector Matching:} 
    The generated vector \( \text{V}_\text{in} \) from Section 3.3.1 is compared with the stored vectors in the Pinecone database using cosine similarity. This process identifies the top-\( k \) most similar results:
    \[
    \text{Top-}k = \{ \text{V}_1, \text{V}_2, \dots, \text{V}_k \},
    \]
    where each \( \text{V}_i \) represents a stored vector that closely matches \( \text{V}_\text{in} \) in the embedding space.

    \item \textbf{Frame Information Retrieval:} 
    Each vector ID in the top-\( k \) results is used to query Neo4j. Neo4j retrieves detailed metadata \( \mathcal{M}_i \) for each matching frame, including file paths, subtitles, and timestamps:
    \[
    \mathcal{M}_i = \{ \text{FilePath}_i, \text{Subtitle}_i, T_{v_i} \}.
    \]

    \item \textbf{Video Clip Assembly:} 
    Using Neo4j's linked-list structure, the system traverses \( S \) frames forward and/or backward from the retrieved frame node to construct a video clip. The variable \( S \) determines the strength or duration of the clip, enabling flexible retrieval:
    \[
    \text{Clip} = \{ F_{i-S}, \dots, F_i, \dots, F_{i+S} \}.
    \]
\end{enumerate}

This retrieval process efficiently combines vector similarity searches with relational metadata navigation. By leveraging Pinecone for fast vector matching and Neo4j for context-rich frame information, the system retrieves not only the most relevant frames but also semantically connected video segments, ensuring a comprehensive and structured approach to video content retrieval.

\subsubsection{Hot-Plug Summary Stage}

In this stage, the retrieved frames and their associated text information are sent back to the Visual Language Model (VLM). The VLM generates a contextual summary of the video clip by integrating the visual and textual data. This summary provides a concise and coherent understanding of the retrieved content, effectively capturing the key details and context of the video segment.

\section{Experiments}
\label{sec:experiments}

\subsection{Single video retrieval experiment}
\begin{figure}
    \centering
    \includegraphics[width=1\linewidth]{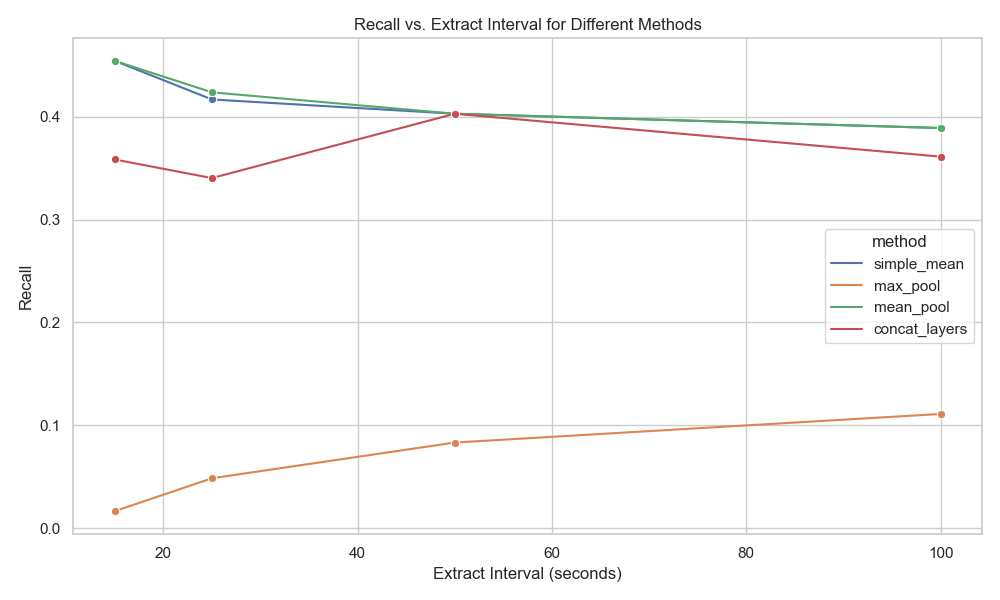}
    \caption{Recall vs Extract Interval and Embedding Method}
    \label{fig:RIEBDM}
\end{figure}

To evaluate the system's performance in single video retrieval (finite space retrieval), we use the target video \cite{redhen_video}. Tests are conducted with extract intervals \( V \) set to 15, 25, 50, and 100, and four embedding methods. Strict conditions are applied, with \( S = 1 \) and \( \text{top}_k = 1 \). For the test samples, frames are extracted at intervals of \(\frac{V}{2}\). A result is considered correct if the test sample's timestamp lies within the retrieval result's time span and is not recorded in the system database. For testing the method "Concatenation of Multiple Layers," we select the last 4 hidden layers as the target. The evaluation metric used in this study is recall, defined as:
\[
\text{Recall} = \frac{\text{Number of Correctly Retrieved result span}}{\text{Total Number of Frames}}
\]
The best results are presented in Table \ref{tab:best_method_results} and Figure \ref{fig:RIEBDM}.
\begin{table}[h!]
\centering
\caption{Best Results for Each Embedding Method Across Extract Intervals}
\begin{tabular}{|c|c|c|c|c|}
\hline
\textbf{Method}       & \textbf{Interval} & \textbf{Correct} & \textbf{Total} & \textbf{Recall} \\ \hline
Simple Mean           & 15                        & 109              & 240            & 0.4542          \\ \hline
Max Pool              & 100                       & 4                & 36             & 0.1111          \\ \hline
Mean Pool             & 25                        & 61               & 144            & 0.4236          \\ \hline
Concat Layers         & 15                        & 86               & 240            & 0.3583          \\ \hline
\end{tabular}
\label{tab:best_method_results}
\end{table}
After inspection, we found that the "Simple Mean" and "Mean Pooling with Attention Mask" methods produce similarly results because the input attention mask consists entirely of ones. Under strict conditions, the system achieves the best result at an extract interval of 15 seconds using the Simple Mean embedding method. This demonstrates the system’s capability to retrieve any frame within a one-hour video with a 30-second margin of error. Furthermore, the performance does not significantly degrade with an increase in the extraction interval, providing evidence of the Vision-Language Model’s (VLM) capability to maintain a continuous hypothesis space for video frame understanding. This capability is critical for effective retrieval tasks. Next, we evaluate the system under looser retrieval conditions by varying \( S \) within the set \(\{1, 2, 3\}\) while keeping \( \text{top}_k = 1 \). The results are presented in Table \ref{tab:Strength_metrics}.

\begin{table}[h]
    \centering
    \caption{Performance Metrics by Strength}
    \label{tab:Strength_metrics}
    \begin{tabular}{|c|c|c|c|c|}
        \hline
        \textbf{Strength} & \textbf{Result Top-k} & \textbf{Recall} & \textbf{Correct} & \textbf{Total} \\
        \hline
        1 & 1 & 0.4542 & 109 & 240 \\
        2 & 1 & 0.5917 & 142 & 240 \\
        3 & 1 & 0.6125 & 147 & 240 \\
        \hline
    \end{tabular}
\end{table}
Under looser retrieval conditions, our system demonstrates a progressive increase in recall rate. At a strength level of 3, the system achieves a 61.3\% recall rate for retrieving the correct 90-second clip based on any provided input image within a one-hour video.
\subsection{Cross video retrieval experiment}
In this section, we will conduct experiments on a cross-video database. This means the system is required not only to return the correct start and end times of the matched content, but also accurately identify the corresponding video from which the frames were extracted. To evaluate the system's performance, we will test it on database with varying numbers of videos within the set \(\{10,25,50\}\) and asses the result using different metrics.
\subsubsection{Retrieve internal video frame}
This task involves enabling users to retrieve specific internal video frames from an existing video dataset. The system processes the query and returns the top-k results, including details such as the video name, start time, end time, and a summary of the relevant video content.

In the evaluation phase, a random video frame will be selected and used as input to test the system's performance. Metrics such as Hit@k and MRR(Mean Reciprocal Rank) will be calculated to measure the accuracy and effectiveness of the retrieval system. The Hit@k is defined as: \[
\text{hit@k}(q) = 
\begin{cases}
1 & \text{If there are matching items in first $k$ hits}\\[6pt]
0 & \text{else}
\end{cases}
\]
for one query, and for N queries, the average value of Hit@K is defined as:
\[
\text{Hit@K} = \frac{1}{N}\sum_{q=1}^{N}\text{hit@k}(q)
\]
And MMR(Mean Reciprocal Rank) is a commonly indicator in information retrieval and recommendation systems, measuring the average position of the correct results returned by the retrieval system on multiple queries.
\[
\text{MRR} = \frac{1}{N} \sum_{q=1}^{N} \frac{1}{r_q}
\] 
where $r_q$ is the rank of the first relevant result for query q and N is the total queries number.

This process not only assesses the system's ability to accurately match the query frame with the correct video and time interval but also evaluates its capability to provide meaningful summaries, which are crucial for enhancing user experience in large-scale video data retrieval. By using a variety of metrics, the evaluation ensures a comprehensive understanding of the system's strengths and areas for improvement.

Therefore, in the cross-video retrieval experiment, we evaluated the system under various search ranges \(\{10,25,50\}\) and conducted experiments in two scenarios: first, retrieving only the correct video; and second, retrieving both the correct video and its corresponding segment. To conduct the experiment, we randomly extracted multiple frames from the existing dataset, performed the retrieval, and then computed the corresponding metrics. Below are the results \ref{tab:onlyvidoeresults} focusing on the scenario where we only retrieve the correct video.
\begin{table}[h!]
\centering
\begin{tabular}{lccc}
\hline
search range & 10    & 25    & 50    \\ \hline
Hit@1     & 0.67  & 0.60  & 0.48  \\
Hit@3     & 0.780 & 0.700 & 0.520 \\
Hit@5     & 0.85  & 0.70  & 0.60  \\
Mean MRR  & 0.858 & 0.924 & 0.858 \\ \hline
\end{tabular}
\caption{Performance metrics with only retrieving the videos}
\label{tab:onlyvidoeresults}
\end{table}

As shown as above, the performance declines when the search range increases from 10 to 50, this pattern is expected: with fewer candidate videos, it’s easier for the system to place the correct one near the top. As the pool grows larger, ranking the correct item among the very top results becomes more challenging.

Below are the results \ref{tab:videoandtimeresult} of retrieving both the correct video and the corresponding time segment.

\begin{table}[h!]
\centering
\begin{tabular}{lccc}
\hline
search range & 10    & 25    & 50    \\ \hline
Hit@1     & 0.29  & 0.30  & 0.24  \\
Hit@3     & 0.430 & 0.420 & 0.280 \\
Hit@5     & 0.49  & 0.42  & 0.34  \\
Mean MRR  & 0.744 & 0.841 & 0.790 \\ \hline
\end{tabular}
\caption{Performance metrics with segment retrieval included}
\label{tab:videoandtimeresult}
\end{table}

As we can see, for Hit@3 and Hit@5, the best performance appears with a smaller search range (10 videos), as expected, since fewer candidates typically mean it's easier to find the correct one within the top few results. Performance generally declines as the search space grows (except for some minor variation at Hit@3 when going from 10 to 25).

According to the both results presented above, we observe that the system’s performance shows a clear improvement when we limit the requirement to retrieving only the correct video, rather than identifying the precise segment within it. This observation suggests that, while our current approach demonstrates reasonable retrieval capabilities, it still has considerable potential for enhancement, especially in terms of fine-grained accuracy.
It’s also worth noting that our evaluation criteria are deliberately stringent. We insist that the system must recall the exact original video from the database, even though different videos may contain highly similar or identical content. This constraint can lower the system’s measured performance, as it punishes cases where the correct content is found but resides in a different video. However, in many practical applications, the goal is simply to discover relevant videos rather than pinpoint a specific, pre-determined source. In such real-world scenarios, where users are satisfied as long as the system returns suitable and contextually related videos, the observed performance would likely be more favorable.

\subsection{Summarization Capability}

\begin{figure}
    \centering
    \includegraphics[width=0.5\linewidth]{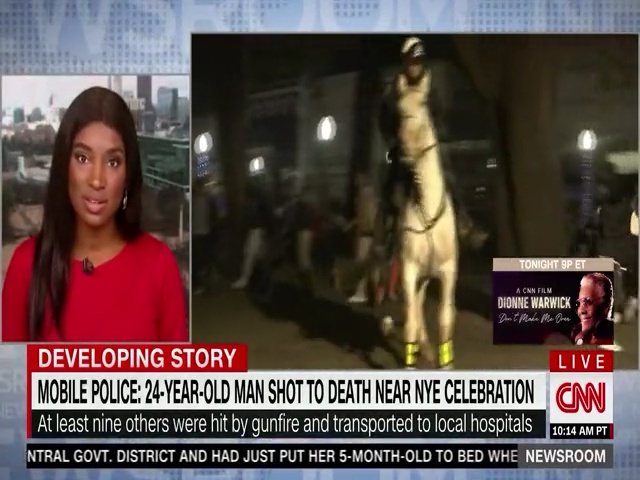}
    \caption{The frame at 900 seconds into the video, the system responds as follows: "The image shows a news segment from CNN featuring a female reporter reporting on a tragic event. The reporter is discussing a shooting that occurred during a New Year's Eve celebration in Mobile, Alabama. The shooting resulted in one person being killed and at least nine others being injured. The reporter is also reporting on a separate story about a machete attack near Times Square in New York City, where a mayor has commended the response to the incident. The segment also includes footage of flooding in California caused by a drenching storm." }
    \label{fig:summarization_example}
\end{figure}

We used the frame at 900 seconds in the video as an example to demonstrate the summarization capability of our system. We employed an extraction interval of 50 seconds and retrieved the top 1 result with a strength level of 2. The corresponding image and summary response are shown in Figure \ref{fig:summarization_example}.

\subsubsection*{Code Availability}
\label{sec:github}
The code and additional resources for this project are publicly available at \url{https://github.com/YichengDuan/svrllm}.
{
    \small
    \bibliographystyle{ieeenat_fullname}
    \bibliography{main}
}


\end{document}